\documentclass[11pt]{article}
\pdfoutput=1
\usepackage{ranlp2011}
\usepackage{times}
\usepackage{url}
\usepackage{latexsym}
\usepackage[utf8]{inputenc}
\usepackage{textcomp}

\title{Considering a resource-light approach to learning verb valencies}

\author{Alex Rudnick\\
  School of Informatics and Computing, Indiana University \\
  Bloomington, Indiana, USA\\
  {\tt alexr@cs.indiana.edu} }

\date{}

\begin{document}
\maketitle

\begin{abstract}
Here we describe work on learning the subcategories of verbs in a
morphologically rich language using only minimal linguistic resources.  Our
goal is to learn verb subcategorizations for Quechua, an under-resourced
morphologically rich language, from an unannotated corpus.  We compare results
from applying this approach to an unannotated Arabic corpus with those achieved
by processing the same text in treebank form. The original plan was to use only
a morphological analyzer and an unannotated corpus, but experiments suggest
that this approach by itself will not be effective for learning the
combinatorial potential of Arabic verbs in general.  The lower bound on
resources for acquiring this information is somewhat higher, apparently
requiring a a part-of-speech tagger and chunker for most languages, and a
morphological disambiguater for Arabic.
\end{abstract}

\section{Introduction}
When constructing NLP systems for a new language, we often want to know the
valence of its verbs, which is to say how many and which types of arguments
each verb may combine with.  Some dictionaries may provide this information,
but even assuming a broad-coverage machine-readable dictionary exists a given
language, that dictionary may not say whether arguments are optional for a
given verb, or how likely they are to occur.

Knowing the selectional preferences and requirements of verbs is useful for
systems that have explicit lexicalized grammars of the languages they cover,
whether for parsing, generation, or both \cite{Briscoe:1997:AES:974557.974609},
and of linguistic interest on its own \cite{gahl}.  The aim of this work was to
build resources for use in L3 \cite{gasser:sxdg}, a rule-based machine
translation system based on dependency grammars, which records the
combinatorial possibilities for every word in its lexicon, and during parsing
and generation constructs a graph describing the structure of the input and
output sentences. We are particularly interested in linguistic resources for
Quechua, which is spoken by roughly 10 million people in the Andean region of
South America, and is thus the largest indigenous language of the Americas.
However, evaluating the approach for Quechua is difficult, due to a lack of
existing lexica and treebanks, so initial experiments have been carried out
with Arabic.

An empirical approach based on a corpus or treebank allows us to learn the
relative frequency with which a given verb takes specific types of arguments.
As a simple example from English, we would like to be able to learn that while
``eat" usually has a direct object, ``put" nearly always has one. Verbs may
also occur with clausal dependents in various ways. For some examples in
English, see Figure \ref{figure1}.

In order to automatically learn this information for resource-scarce,
morphologically rich languages, we set out to implement a system that requires
only an unannotated corpus and a morphological analyzer; other recent
approaches have made use of more linguistic knowledge, in the form of
treebanks, parsers, or chunkers. In practice, our we will also require more
resources to be fruitful; this may be addressed in the future.

\begin{figure*}
\begin{center}
\begin{tabular}{|l|l|}
 \hline 
 {* I put.}                       &  {I believe that he is tall.    } \\
 {* I put the potato.           } &  {I consider him tall.          } \\
 {I put the potato on the table.} &  {* I consider that he is tall. } \\
 \hline
\end{tabular}
\end{center}
\caption{``What did you do yesterday?", and \emph{believe} vs. \emph{consider}}
\label{figure1}
\end{figure*}

\section{Related work}
Many other researchers have addressed the problem of documenting the properties
of the different verbs in a given language, using evidence from corpora and
manual lexicography. Automatic approaches have the potential to involve
less manual work avoid human biases, giving a more objective measure of the
behavior of a given verb.

We see in the literature a few different terms that describe the combinatorial
potential of a verb, including \emph{subcategorization},
\emph{subcategorization frames}, \emph{valence} or \emph{valency}. In any case,
these terms describe which arguments and adjuncts may appear with a given verb,
how often, and which ones are obligatory. 
While describing similar notions, these terms do not seem to be
interchangeable; while this work is concerned with with ``surface level"
syntax, looking for arguments that are present in practice (such that a parser
could find them), in Functional Generative Description (FGD), the term valency
refers to a tectogrammatical notion; arguments might be known to the speaker
but not expressed. This deeper notion cannot be readily observed from text alone, as
pointed out by Bojar \shortcite{bojar:2003}.

\subsection{Valency lexica for English}
Gahl \textit{et al.} \shortcite{gahl} describe a study in which they had a team
of linguists annotate thousands of English sentences -- 200 sentences for each
of 281 English verbs of interest -- and build a table of distributions of
subcategorization frames that they observed for each of the verbs.  They
describe the difficulties that may be faced in trying to learn
subcategorizations from a corpus: in a given body of text (even one as big as
the Brown corpus), it may be that not all possible subcategorizations will be
observed. Additionally, different genres of text may exhibit different verb
usage. This paper also gives a good overview of the uses of valency information
and a view on verb subcategorization from psycholinguistics, including
elicitation experiments that psycholinguists have used to learn the relative
frequencies of different uses of verbs.

Gahl's group has made their results available in machine-readable form,
providing a potentially useful resource for those interested in English verbs.
However, their approach was very labor-intensive and required a large corpus.

Ushioda \textit{et al.} \shortcite{ushioda} describe earlier work on acquiring verb
subcategorizations for English. Their method requires a tagged
corpus, although an untagged corpus and an accurate tagger would work as well.
On the basis of the tags, they perform partial parsing to identify noun phrases
(chunking), and then use some simple rules specified in terms of regular
grammars to identify common patterns of constituents in the sentences, which
are marked with corresponding subcategorization frames. This approach does not
require the use of a deep parser, but the rules had to be crafted specifically
for English. 

Ushioda explores the WSJ corpus with this extractor, and reports results on 33
randomly selected common verbs: the extraction rules achieve 86\% accuracy over
sentences from a test set taken from WSJ, where the correct subcategorization
frames for the test set had been determined manually.

Brent \shortcite{DBLP:journals/coling/Brent94} addresses the seeming impasse that in
order to get accurate parses automatically, one needs to know about the
syntactic frames of different verbs, but in order to get the frame information
from a corpus, the sentences must be parsed accurately. He handles this problem
by crafting language-specific rules that initially only refer to closed-class
words and do not require complete parses. This approach is somewhere between
having no syntactic knowledge at all, and requiring a large grammar of the
target language: to start out, it must first figure out which words in the
corpus are verbs. He then uses statistics to infer previously unknown facts
about the language, for example, which English verbs can occur with each of six
different kinds of arguments.

While this appears to be an effective approach, one wonders how hard it would
be to apply to an unfamiliar language. Producing the initial rules may require
a lot of linguistic insight; for example, Brent relies on the fact that in
English, verbs typically do not appear immediately after determiners or
prepositions. In a language with more free word order, or one without
determiners or prepositions, what sorts of rules might one use?

Briscoe and Carroll \shortcite{Briscoe:1997:AES:974557.974609} describe a system
that finds subcategorization frames for verbs in English, including relative
frequencies for each class for a given verb. They adopt a very detailed scheme
for verb subcategories, in which each usage falls into one of 160 different
classes, where each class includes specific information about particles and
control of the arguments of the verb. Their system requires the use of a POS
tagger, a lemmatizer, and a pre-trained probabilistic parser; after identifying
and classifying the different subcategories of verb usages, they incorporate
this information into another parser and demonstrate that it improves parsing
accuracy.

\subsection{Valency lexica for Slavic languages}
The VALLEX project \cite{zabortsky} has produced a large hand-curated database
of valency frames for verbs in Czech, covering roughly the 2500 most common
verbs in Czech and cataloging their various senses. VALLEX makes use of
Functional Generative Description as the background linguistic theory for its
account of verbs, and so records, at least, whether a verb sense takes an
\emph{actor}, \emph{addressee}, \emph{patient}, \emph{effect}, and
\emph{origin}, and whether these must be specified, as well as a large number
of other ``quasi-valency complementations" and ``free modifications". VALLEX
provides a very detailed account of the potential uses of each verb in its
lexicon, much more detailed than what can currently be produced with automatic
methods.

More recently, Przepiòrkowski has done work focusing on Polish, comparing
valence dictionaries built with the use of shallow parsing to those built with
deep parsing  \shortcite{przepiorkowski}. Because his shallow parser may not handle
all of the sentences in the corpus, his approach ends up ignoring more than
half of the training data, but from the remaining 41\% of the IPI PAN Corpus,
he collects counts of the different frames in which each verb was observed, and
uses a small number of Polish-specific rules to post-process the observations,
then does statistical filtering to try to reduce noisy observations.

Przepiòrkowski evaluates the extracted lexical information in two different
ways, making use of both pre-existing valence dictionaries and sentences
hand-annotated by linguists, finding that his shallow-parsing technique
actually produces results that agree more closely with frames that were
observed in the texts by linguists than the existing valence dictionaries.

Debowski \shortcite{DBLP:journals/lre/Debowski09} presents a procedure for
extracting valence information and frame weights for Polish that makes use of a
non-probabilistic deep parser and a novel use of EM, which he says is simpler
than the more traditional repeated inside-outside approach to optimizing
weights for a probabilistic grammar. Additionally, in his EM formulation, the
weight-optimization problem is convex, so he can start with uniform prior
probabilities and be guaranteed to get a globally optimum solution. Debowski
also includes an approach for filtering incorrect frames that were found in the
parsed text.

When analyzing his results, Debowski notes that some of his observed ``false
positives" described valid uses of the verbs in question, but were not
included in the compiled valence dictionaries that he used in evaluating his
approach.

\subsection{Valency Lexica for Arabic}
Informed by the Prague Arabic Dependency Treebank and the Functional Generative
Description (FGD) theory of syntax, Bielický and Smrž \shortcite{BIELICK08.578}
describe desiderata for a valency lexicon for Arabic. They do not describe the
production of such a lexicon in practice, but lay out a framework for
discussing one, proposing a structure for lexical entries in the valence
dictionary. Their structure is based on VALLEX, which seems to have a broadly
applicable formalism for describing verbs.  They also describe some tools
useful for the task, including an FST-based morphological analyzer for Arabic,
and explain FGD's account of verbal arguments/adjuncts.

\subsection{Resources for Quechua}
Rios \textit{et al.} \shortcite{rios} address the more general problem of
acquiring enough linguistic knowledge to build effective NLP systems for
under-resourced languages such as Quechua, with a more labor-intensive
approach. They describe their construction of a phrase-aligned treebank for
Quechua and Spanish, which covers about 200 sentences, with text from the
\emph{Declaration of Human Rights} (available in many languages, including
Spanish and Quechua) and the website of \emph{La Defensoría del Pueblo}, a
Peruvian government organization that advocates for citizens rights. Aside from
the morphological analysis of Quechua, the treebanking and alignment process
currently require human attention, though this may be partially automated in
the future.

The treebank so far is small, but it may be increasingly useful for machine
translation as their treebanking process becomes more automated. Rios \emph{et
al.} note a surprising number of available bitexts for Spanish/Quechua,
including political texts, news, translated novels and poetry.  

\section{Proposed Approach}
Our approach starts by processing each sentence in the corpus with the
morphological analyzer, thus finding all of the verbs. For sentences with only
one verb, we then count the occurrences of nouns that seem to be, because of
inflection, the arguments of the verb. Here plausible verb arguments will need
to be identified with a small number of language-specific heuristics. For
example, a noun inflected with the accusative case in a sentence with a verb
and a clear subject will likely be the object of that verb. This approach
throws away the information from sentences with multiple verbs and embedded
clauses, but it does not require syntactic analysis. We had hoped that the
frequencies learned with this approach will approximate the frequencies that
would be learned using deeper syntactic analysis, but this does not bear out
empirically.

Noisy observations could be filtered out using an approach similar to the one
described in \cite{przepiorkowski}. In the long run, for consistency, we would
like to build a lexicon in the VALLEX style, discovering whether each given
verb usage contains an explicit Actor, Addressee, Patient, Effect, and Origin,
when these roles can be identified by the morphological cues.

\subsection{Evaluating Valency Learning Techniques}
When building a system that builds valency lexica for the verbs of a given
language, we would like both good recall, meaning that the system identifies a
many of the verb usages that are actually present in the training text, and
high precision, meaning that the answers the system returns are actually
correct. To measure both of these, we can take some preexisting lexicon to be
the gold standard, but good valency lexica are not available for most
languages.

What we can do instead is take the verb usages in a treebank, and consider the
subcategorization lexicon constructed in that way to be the gold standard. We
have an Arabic treebank (Arabic Treebank Part 1, v3.0) available from the LDC
\cite{Maamouri05arabictreebank}, so for this work we make use both of that
treebank and the associated flat text. We chose Arabic for its rich morphology,
and for the somewhat convenient, though not freely redistributable, treebank.
If the results were good for Arabic, then that would be evidence that it might
be helpful for constructing valency lexica for other languages as well.

\section{Experiments with Arabic}
We carried out experiments with the text of the Arabic treebank, using both
the transliterated text with syntactic annotations and the unannotated Unicode
text in Arabic script. Given the treebank annotations, we can find the verbs in
each sentence, as well as the other components of the verb phrases, quite
easily by traversing the parse trees. For initial experiments due to the
sparsity of the data, we pass over the problem of deciding whether a
constituent is an argument or adjunct of the verb.

To find all of the verb subcategory frames in the treebank, we traverse the
tree of each sentence and record the immediate children of the verb phrase that
are not the verb itself. These are considered a set, and recorded with the stem
of the verb in question. The process is described in more detail in Figure
\ref{treerules}.

\begin{figure*}
\begin{center}
\begin{itemize}
\item For each sentence in the treebank...
  \begin{itemize}
  \item For each verb phrase in that sentence...
    \begin{itemize}
    \item Look for a word in the VP with a tag that contains one of IV, PV,
    IV\_PASS or PV\_PASS (one may not be present; if so, skip this VP)
    \item Find the stem for the verb, if present
    \item Record the verb stem, along with the tags of the sibling constituents.
    \end{itemize}
  \end{itemize}
\end{itemize}
\end{center}
\caption{Process for finding verbs and arguments in the treebank}
\label{treerules}
\end{figure*}

Considering the entire treebank, which consists of 734 news documents, there
are 5845 sentences, containing 14115 verb phrases. The majority (92\%) of these
verb phrases have a verb that can be found with the rules described. The most
common verb stems, presented in Buckwalter transliteration, were: ``kAn",
``qAl", ``\textgreater{}aEolan", ``\textgreater{}aDAf", ``\textgreater{}ak~ad", ``kuwn", ``\textgreater{}awoDaH", ``*akar",
``mokin", ``\textgreater{}afAd".
Each of these occurred at least 100 times in the corpus.
Not all of these can be translated sensibly to English without context by
Google Translate, but using it as a glossing tool, we get: \emph{was},
\emph{declared}, \emph{added}, \emph{confirmed}, \emph{fact}, \emph{clear},
\emph{enabled}, and \emph{reported}.  There were 1747 different verb stems
observed altogether.

Adapting the approach of Przepiòrkowski \shortcite{przepiorkowski}, we then
focus on the sentences from the corpus that contain only one verb. This
allows us to avoid making attachment decisions, since deep parsers may not be
available or reliable. Bojar \shortcite{bojar:2003} does something similar,
with the addition of a chunker that can find subordinate clause boundaries.
This approach also seems sensible particularly for Quechua and Arabic, since
case is typically marked on nouns for both languages, although this still
leaves the problem of dropped arguments.

On its own, filtering out a large number of sentences is not a problem;
that we include a nontrivial fraction of the sentences at all is promising.
To improve coverage, we could simply feed the system more unannotated text. For
Arabic, we could use the very large supply of Arabic news available on the web.
This approach would be less plausible for Quechua, although of course
unannotated Quechua text is more plentiful than Quechua treebanks.

\begin{figure*}
\begin{center}
\begin{tabular}{|l|l|l|}
 \hline 
                                  &  count          & fraction \\ \hline
  sentences in PATB part 1        &  5845 & 1.0  \\ \hline
  sentences with only one VP      &  926  & 0.16 \\ \hline
  unique verb stems observed      &  1747 & 1.0 \\ \hline
  unique verb stems in sentences with only one VP & 376 & 0.22 \\ \hline
\end{tabular}
\end{center}
\caption{Filtering results on sentences from the Penn Arabic Treebank}
\label{figure2}
\end{figure*}

\subsection{Sentence Selection in Practice}
We might wonder, however, whether our sampling of sentences leads to biases in
the observed verbs and their usages.  Considering English verbs such as
``think", ``believe", or ``request", which usually occur with some clausal
argument that includes some other verb, we imagine that the analogous verbs in
the language we are investigating would be under-represented or simply not
learned at all.

Experiments showed that both of these worries are well-founded: sentences that
had only one verb had a substantially different distribution of verb stems. The
verbs that were most common in the one-verb sentences were ``saj~al", ``qAl",
``\textgreater{}aEolan", ``\$Arik", ``kAn", ``daEA", ``fAz", ``balag",
``\textgreater{}aHoraz", and ``lotaqiy".  These are glossed by Google
Translate as: \emph{record}, \emph{said}, \emph{announced}, \emph{was},
\emph{called}, \emph{beat}, \emph{was}, \emph{made}, and \emph{assess},
definitely a different sort of verb than the ones we see commonly in the in the
text generally.

Even among the verbs that happen commonly in both the text in general and the
one-verb sentences, we observe different usages. The most common verb in
general, ``kAn" (\emph{was}), occurs with another verb phrase as an argument
about 400 times in the treebank. The second-most common verb in general and in
the one-verb sentences, ``qAl" (``declared/said"), most often occurs with an
SBAR (indicating a nested clause) in general, but of course these usages do not
occur in the one-verb sentences.

\subsection{Morphological ambiguity}
In experiments with the nearly-unannotated\footnote{very lightly marked-up with
SGML} text distributed with the treebank, we made use of AraMorph
\cite{brihaye}, a Free Software version of the Buckwalter morphological
analyzer that handles Unicode text.  The goal with the unannotated text was to
see which subcategory frames we could observe in sentences with a single verb
-- the rich Arabic morphology usually marks case on nouns, which should allow
us to find many arguments to verbs.  We could then compare the valencies
learned from the unannotated corpus with those that are more easily observable
from the treebank.  If the valencies that we discover with the unannotated
approach are close to those learned from the treebank, and we get a broad
coverage over the verbs observed in the corpus, then this would provide an
argument that the technique works fairly well for Arabic, and we could continue
using it as we acquire more textual data for more under-resourced languages. 

However, Buckwalter-style morphological analyzers do not account for
morphological ambiguity, which would present difficulties in the long run. This
problem is particularly dire because the Arabic script is an \emph{abjad},
which is to say that it only records the consonants for each word. There is an
optional system for annotating the vowel sounds as well, but it is often not
used in practice.  This presents a problem for Arabic-language NLP systems,
though, since a word without context rarely has a unique morphological
analysis. In fact, within the corpus, we observed a mean of about 7.5 possible
Buckwalter analyses per word, with $\sigma = 8.4$, and a maximum of 86.
\footnote{The word with 86 Buckwalter analyses can mean, at least, ``one",
``and scrutinize", ``and sharpen", or ``and be furious".}

We also briefly experimented with a Quechua morphological analyzer, Michael
Gasser's {\tt AntiMorfo} system \shortcite{gasser:antimorfo}; it can analyze Quechua
verbs, nouns, and adjectives.  We also see morphological ambiguity in Quechua
words, although it is not nearly so striking, largely due to the orthography
with vowels. We saw a mean of 1.7 analyses per word, with $\sigma = 1.3$, and a
maximum of 10. For Quechua, we used a small corpus produced by the AVENUE
project, described in \cite{monson06}, which includes bitext elicited from
native speakers, and monolingual text, both from UN documents and local
stories. Interestingly, with the FST-based morphological analysis, we cannot
tell whether a word is definitely a verb. For example, \emph{waqaychu} may be a
verb, noun, or infinitive, as ``waqa" is in AntiMorfo's lexicon as a verb root,
and ``waqay" as a noun root. So to get good results for Quechua, we would need
a POS tagger or other means to choose between analyses.  As far as treebanks of
Quechua, for evaluating an automatically-extracted Quechua verb lexicon, to our
knowledge there is no large one available, although Rios \textit{et al.} are
developing a small one \shortcite{rios}.

\section{Conclusions and future work}
While the problem of discovering grammatical subcategories of verbs remains
interesting, and solutions to that problem would be of practical use, it is
almost definitely not enough to use only a morphological analyzer and a
medium-sized unannotated corpus for this purpose; disregarding sentences with
more than one verb leads to misleading view of the language as a whole, and
selection preferences learned from these sentences would not be suitable for
parsing most sentences. To make better use of the existing data, it would be
helpful to have a chunker to find the boundaries of clauses and noun phrases,
as in \cite{przepiorkowski}. For Czech verbs, Bojar similarly used finite-state
rules to find coordinated and subordinate clauses \shortcite{bojar:2003}.

While ambiguity in morphological analysis can be a hurdle in any language, the
vowel-free nature of typical Arabic text presents a particularly serious one;
we would like to be able to use any available unannotated text, but without
morphological disambiguation, we are left with many possible interpretations
for most tokens. This could be mitigated with software like MADA+TOKAN
\shortcite{habash:2009}, which chooses the most likely morphological analysis for a
given context; for other languages, such as Quechua, similar tools will need to
be developed.

Even with proper POS tagging, morphological disambiguation, and chunking, we
are still faced with the problem of negative evidence; without a very large
corpus, we cannot say with confidence that a given verb cannot appear with
certain arguments, simply because it has not yet been observed with those
arguments. This difficulty suggests an active learning approach, perhaps
coupled with crowdsourcing. We could imagine a system that generates sentences
to test the hypothesis that a given verb may be used with a given
subcategorization frame, then presents those sentences to human users for
grammaticality judgments.
\bibliographystyle{acl} \bibliography{alexr-valency.bib}
\end{document}